\documentclass{anstrans}

\title{Deploying Frontier Agentic Technology in MOOSEnger, a Multiphysics-Capable AI Assistant}
\author{Zaid Abulawi,$^{*}$ Mengnan Li,$^{\dagger}$, Guillaume Giudicelli,$^{\dagger}$, Yang Liu,$^{*}$, and Cody Permann,$^{\dagger}$}

\institute{
$^{*}$Department of Nuclear Engineering, Texas A\&M University (TAMU), TX, USA, zaidabulawi@tamu.edu \& y-liu@tamu.edu
\and
$^{\dagger}$Idaho National Laboratory (INL), ID, USA, mengnan.li@inl.gov \& Guillaume.Giudicelli@inl.gov
}

\usepackage{graphicx} 
\usepackage{booktabs} 
\usepackage{microtype} 
\usepackage{tikz}
\usepackage{xcolor} 
\usetikzlibrary{arrows.meta,positioning,shapes.geometric}
\usepackage{fontawesome5}

\usepackage{booktabs}    
\usepackage{makecell}      
\usepackage{glossaries}
\usepackage{tabularx}
\makeglossaries

\newacronym{moose}{MOOSE}{Multiphysics Object-Oriented Simulation Environment}
\newacronym{moosenger}{MOOSEnger}{MOOSEnger}

\begin{document}
\section{Introduction}

\begin{figure*}[h]
    \centering
\includegraphics[width=1\linewidth]{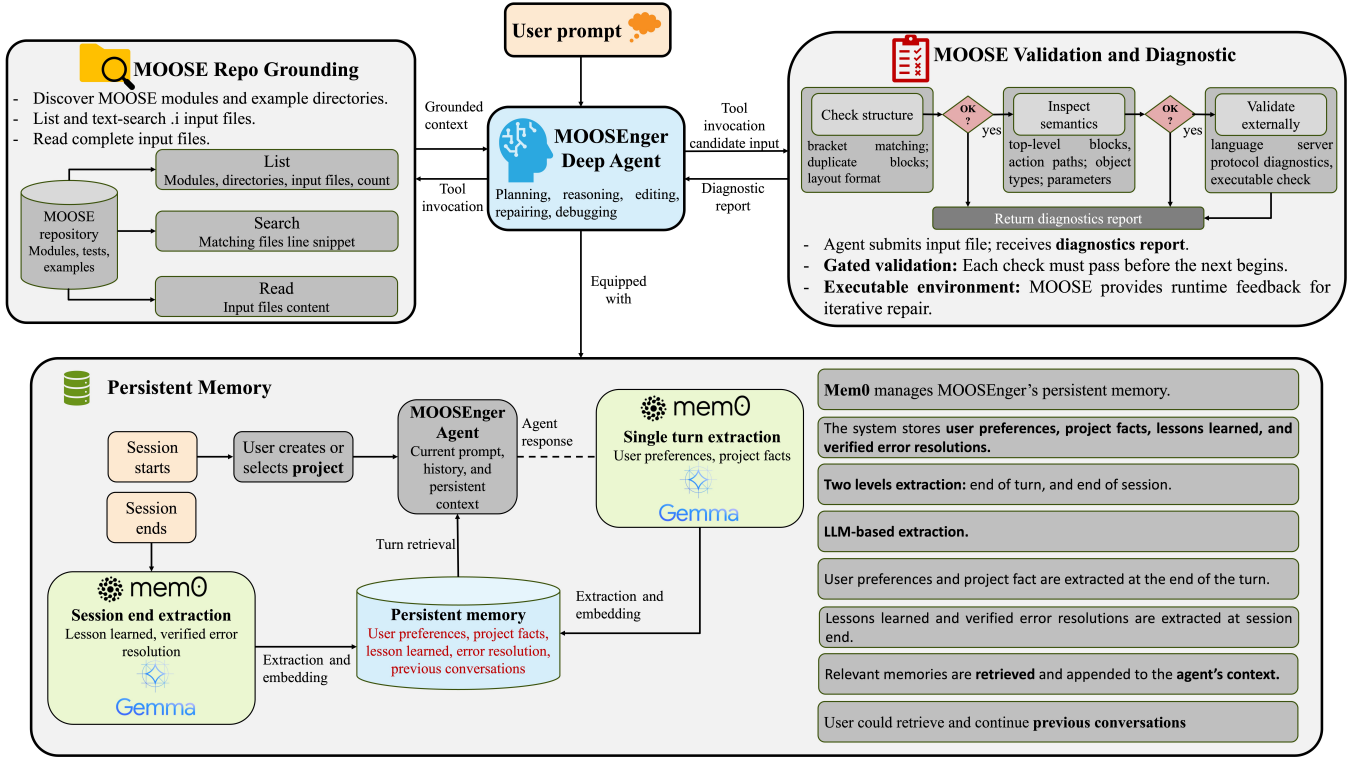}
    \caption{MOOSEnger methodology for evaluating locally-hosted models and
improving MOOSE-aware context engineering through Retrieval / Knowledge
Grounding, environment interaction for validation and repair, and persistent
memory.}
    \label{fig:methodology}
\end{figure*}

The Multiphysics Object-Oriented Simulation Environment (MOOSE)~\cite{HARBOUR2025102264} is an open-source finite-element framework for building multiphysics simulation applications. Using a multiphysics environment effectively demands specialized expertise, creating a barrier for many domain scientists and engineers. Domain-specific knowledge and tailored tools substantially enhance a Large Language Model's (LLM) performance on demanding tasks like modeling and simulation. A specialized agent, rather than a generic LLM agent, can therefore better support experts, close the expertise gap, and deliver impact across a broad community of MOOSE users.

MOOSEnger, developed at Idaho National Laboratory (INL), is a domain-specific, tool-enabled AI agent built for the MOOSE Framework~\cite{li2026moosenger}. Its deep-agent workflow pairs language-model reasoning with simulation actions: the agent retrieves relevant MOOSE documentation and examples, drafts or edits input files, and calls MOOSE-aware tools to check, and run the resulting simulations.

Simulations using MOOSE-based applications are set up using Hierarchical Input Text (HIT) files, which specify the mesh, variables, materials, boundary conditions, execution settings, and outputs. LLMs have shown strong performance on text-based and coding tasks. Modeling and simulation applications share some challenges with other code-generation domains, but also introduce a distinct one. Like structured code, HIT input files must follow strict formatting and use parameter relations; parameters within a single file are tightly coupled, so changes must be propagated. However, they are distinct in two ways. First, despite being open source, its solvers were likely not part of the fine-tuning data. Second, modeling and simulation inputs require physical understanding: an input's validity becomes clear from runtime diagnostics, and even a simulation that runs successfully may be physically wrong.

This work extends MOOSEnger with a harness focused on locally-hosted, open-weight language models. Locally-hosted models mitigate the risk of data leakage, as MOOSE is frequently applied to export-controlled and sensitive data. The harness gives the agent a full pipeline: it retrieves contextual knowledge from the MOOSE repository, validates and diagnoses the resulting input through interaction with the simulation executable environment, and extracts and stores lessons in a persistent memory for reuse across sessions.

The resulting framework is demonstrated on an engineering problem from the National Reactor Innovation Center Virtual Test Bed (VTB), illustrating its potential to support realistic multiphysics simulation workflows~\cite{wozniak2021duct,giudicelli2023vtb}. Additionally, the agent performance is evaluated on different categories including diffusion, Navier--Stokes, phase field, plasticity, porous media flow, solid mechanics, transient heat transfer, and reactor mesh generation. Each category consists of 25 prompts/cases. We compare MOOSEnger-Gemma4 against MOOSEnger-GPT-5.2, alongside baseline Gemma4 and GPT-5.2 without agentic capabilities. MOOSEnger-GPT-5.2 shows a slight edge, achieving a 90\% success rate versus 76.5\% for MOOSEnger-Gemma4. The baseline models perform far worse, at just 5\% (GPT-5.2) and 0\% (Gemma4), underscoring the impact of the agentic harness.

\section{Methodology}

The methodology equips the MOOSEnger deep agent with tools for grounding and retrieving relevant input files and examples, running structured diagnostics through an interactive execution environment, receiving feedback and repairing the input accordingly, and extracting reusable knowledge into persistent memory. Implemented using LangGraph, the agent follows a stateful reasoning-and-action loop in which it interprets the task, selects and invokes appropriate tools, and evaluates their outputs. Figure~\ref{fig:methodology} summarizes this workflow.

\subsection{MOOSE Repository Grounding}

The first source of task-specific context is the local MOOSE source tree. MOOSE includes a core framework and different physics and numerical modules under the \texttt{modules} directory, including heat transfer, phase field, Navier--Stokes, solid mechanics, porous flow, and reactor meshing. These modules contain tests, examples, and tutorials that demonstrate valid HIT syntax, MOOSE objects, parameters, execution settings, and established modeling practices.

Before drafting an input file, the MOOSEnger deep agent identifies modules relevant to the user’s request and explores their example and test directories. It then reads representative HIT files to find reusable block structures, object configurations, naming conventions, and solver settings. Because these files come from the MOOSE repository, they provide concrete examples grounded in working MOOSE applications.

This functionality allows the agent to list modules, browse directories containing input files, enumerate candidate files, and read a selected file in full. It can also search HIT files for object names, parameters, or other syntax. By default, the search compares each line using a case-insensitive substring match and returns a bounded set of file paths, line numbers, and matching snippets. This helps the agent move from a general modeling request to relevant repository examples.

\subsection{MOOSE Validation and Diagnostic}

After drafting or editing an input file, MOOSEnger checks it through a sequence of deterministic, MOOSE-aware validation steps. It first checks the HIT structure, including bracket balance and duplicate blocks, so malformed files can be reported before more expensive semantic checks are attempted. It then verifies block names, action paths, object types, and parameter names using information from the local MOOSE environment.

Once the file passes these static inspections, MOOSEnger queries the Language Server Protocol (LSP). LSP feedback provides line- and column-level diagnostics, formatting support, symbol context, and parameter-template information that can identify invalid keys or misplaced configuration blocks more precisely than free-form model judgment. The candidate is then checked against the target MOOSE executable using \texttt{--check-input}, which verifies any input file execution-time errors.

The results are combined into a concise diagnostic report containing each issue’s source object, message, location in the input, type, and suggested next step if any. The agent uses this evidence to repair the actual input file and runs the checks again. When execution is requested, the same environment can perform a mesh-only check, a short test run, or a full simulation, helping the agent distinguish input errors from convergence or physics-modeling problems.

\begin{figure*}[!ht]
    \centering
    \includegraphics[width=0.95\textwidth]{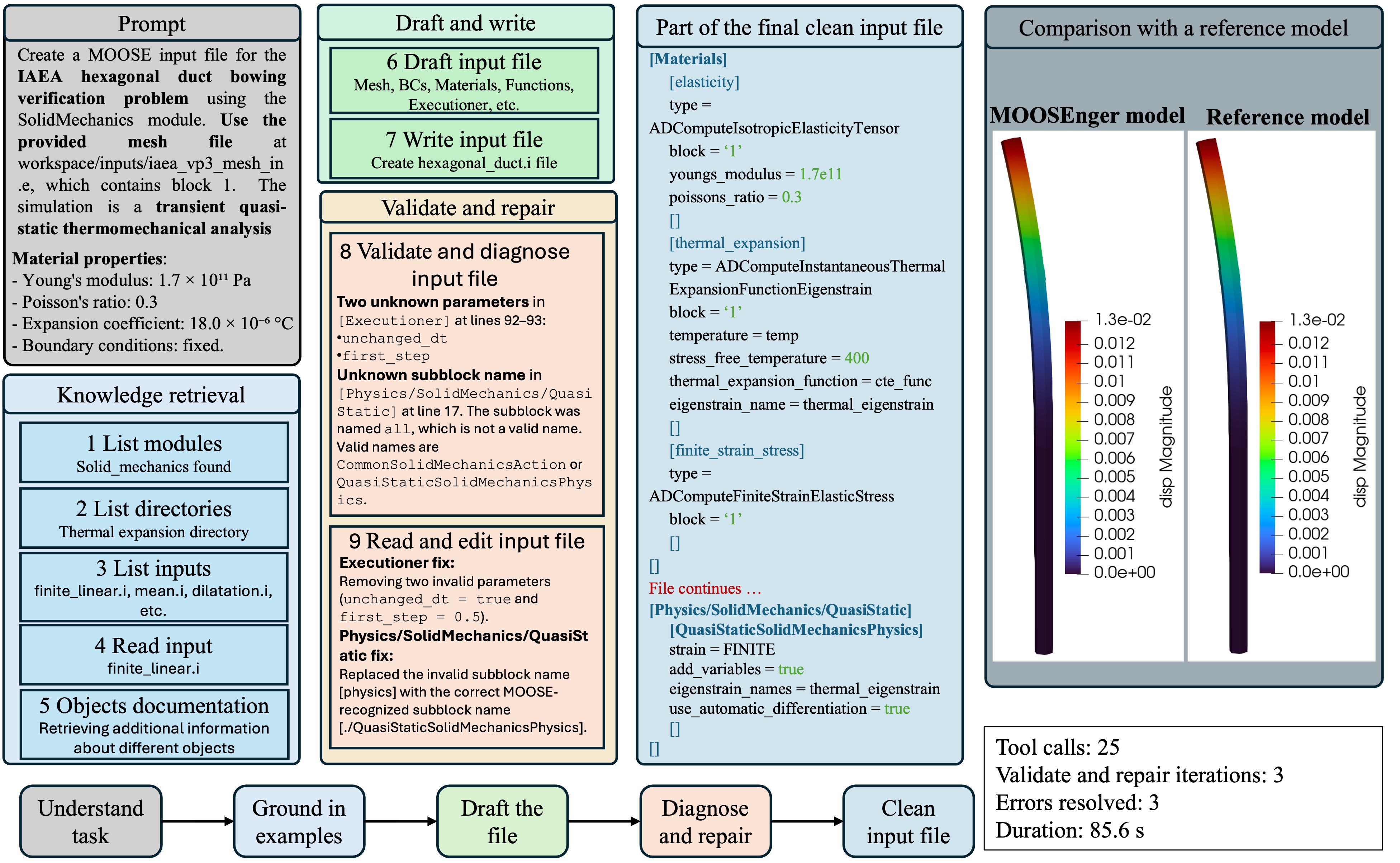}
    \caption{MOOSEnger agent trace for the IAEA hexagonal duct bowing verification problem. The agent follows a five-phase pipeline: 
    (1)~\textbf{understand task} — parse the prompt, including user-input material properties; 
    (2)~\textbf{ground in examples} — retrieve relevant MOOSE modules, directories, and reference inputs; 
    (3)~\textbf{draft the file} — using inference, drawing the specific syntax from the context; 
    (4)~\textbf{diagnose and repair} — run the input precheck tool three 
    times, resolving issues and errors across two edit iterations; and 
    (5)~\textbf{finalize input file} — produce a validated HIT file ready for execution.}
    \label{fig:moosenger_trace}
\end{figure*}

\subsection{Persistent Memory}

MOOSEnger has the capability to restore conversation history, allowing the agent to resume an earlier session with its messages and tool interactions. However, it remains tied to one session and does not organize knowledge learned across multiple tasks. Persistent memory addresses this limitation by building a reusable knowledge database from user–agent interactions across sessions.

MOOSEnger uses Mem0 to manage four memory types: user preferences, project facts, lessons learned, and verified error resolutions. Mem0’s main operations are used to add new memories, and retrieve relevant memories with semantic. A locally hosted Gemma 4 model performs the LLM-based extraction, turning conversation and tool evidence into concise memory records.

LLM-based extraction using Gemma 4 occurs at two points in the workflow. After each turn, a background process uses the LLM to extract user preferences and project facts from the latest interaction. At session end, the LLM extracts broader lessons and verified error resolutions, using validation evidence to avoid storing unsuccessful repairs.

At session startup, a bounded set of user preferences and project facts is loaded into a temporary cache. During each turn, Mem0 searches for lessons and error resolutions relevant to the current prompt. The retrieved memories are then appended to the agent’s context, allowing MOOSEnger to combine the current conversation with useful knowledge from earlier sessions.

\section{Case-Study Demonstration}

The demonstration case uses a sodium fast reactor hexagonal duct bowing model from the VTB. This case is an IAEA benchmark which examines free thermal bowing of a mechanically fixed, three-dimensional hexagonal duct subjected to radial and axial thermal gradients. The published case uses the MOOSE Reactor-module mesh generation and the Solid Mechanics module to calculate thermo-mechanical deformation~\cite{wozniak2021duct}. This case is well suited for demonstration because it exercises MOOSE input features that are challenging for an agent: coupling thermal and mechanical physics, interplay between functions, auxiliary variables, Dirichlet boundary conditions, solid-mechanics physics compact syntax, material models, and post-processing to produce benchmark results.

Figure~\ref{fig:moosenger_trace} shows the resulting agent trace. The workflow begins with task interpretation, then grounds the model by listing MOOSE modules, browsing relevant example directories, listing candidate input files, reading a reference input, and retrieving object documentation. The agent then drafts and writes the input file, validates it, and enters a repair loop in which the diagnostics identify invalid \texttt{Executioner} parameters and an invalid Solid Mechanics subblock name. Across 25 tool calls, the run required three validation-and-repair iterations, resolved three errors, and completed in 85.6\,s. The final clean input is shown beside the reference model to illustrate that the repaired MOOSEnger-generated input files produces the expected duct-bowing deformation pattern.

\begin{table*}[h]
\caption{End-to-end results for the 200-prompt \gls{moose} benchmark: harness-enabled (\gls{moosenger}) vs.\ standalone LLM performance, across GPT~5.2 and Gemma 4~31b backends.}
\label{tab:results_full_comparison}
\centering
\normalsize
\renewcommand{\arraystretch}{1.15}
\begin{tabular}{@{} l c c c c @{}}
\toprule
\textbf{Problem family} &
\makecell{\textbf{\gls{moosenger}}\\\textbf{+ Gemma 4 31b}} &
\makecell{\textbf{\gls{moosenger}}\\\textbf{+ GPT 5.2}} &
\makecell{\textbf{ChatGPT 5.2 API}\\\textbf{standalone}} &
\makecell{\textbf{Gemma 4 31b}\\\textbf{standalone}} \\
\midrule
Diffusion &
\makecell[c]{23/25\\(92\%)} &
\makecell[c]{25/25\\(100\%)} &
\makecell[c]{9/25\\(36\%)} &
\makecell[c]{0/25\\(0\%)} \\
Transient heat conduction &
\makecell[c]{21/25\\(84\%)} &
\makecell[c]{23/25\\(92\%)} &
\makecell[c]{0/25\\(0\%)} &
\makecell[c]{0/25\\(0\%)} \\
Solid mechanics &
\makecell[c]{22/25\\(88\%)} &
\makecell[c]{24/25\\(96\%)} &
\makecell[c]{0/25\\(0\%)} &
\makecell[c]{0/25\\(0\%)} \\
Plasticity &
\makecell[c]{15/25\\(60\%)} &
\makecell[c]{21/25\\(84\%)} &
\makecell[c]{0/25\\(0\%)} &
\makecell[c]{0/25\\(0\%)} \\
Porous flow &
\makecell[c]{17/25\\(68\%)} &
\makecell[c]{23/25\\(92\%)} &
\makecell[c]{1/25\\(4\%)} &
\makecell[c]{0/25\\(0\%)} \\
Navier--Stokes &
\makecell[c]{16/25\\(64\%)} &
\makecell[c]{21/25\\(84\%)} &
\makecell[c]{0/25\\(0\%)} &
\makecell[c]{0/25\\(0\%)} \\
Phase field &
\makecell[c]{17/25\\(68\%)} &
\makecell[c]{21/25\\(84\%)} &
\makecell[c]{0/25\\(0\%)} &
\makecell[c]{0/25\\(0\%)} \\
Reactor mesh generation &
\makecell[c]{22/25\\(88\%)} &
\makecell[c]{21/25\\(84\%)} &
\makecell[c]{0/25\\(0\%)} &
\makecell[c]{0/25\\(0\%)} \\
\midrule
\textbf{Overall} &
\makecell[c]{\textbf{153/200}\\\textbf{(76.5\%)}} &
\makecell[c]{\textbf{179/200}\\\textbf{(90\%)}} &
\makecell[c]{\textbf{10/200}\\\textbf{(5\%)}} &
\makecell[c]{\textbf{0/200}\\\textbf{(0\%)}} \\
\bottomrule
\end{tabular}
\end{table*}

\section{Agent Evaluation}
The evaluation focuses on end-to-end, tool-enabled MOOSE input authoring with the MOOSEnger agent architecture. The benchmark prompts are organized into physics-oriented categories including diffusion, Navier--Stokes, phase field, plasticity, porous media flow, solid mechanics, transient heat transfer, and reactor mesh generation. A prompt is considered successful when the generated input runs successfully and remains aligned with the requested physics, geometry, boundary conditions, and outputs. 

Within the \gls{moosenger} harness, GPT~5.2 resolves 179 of 200 prompts (90\%), compared to 153 of 200 (76.5\%) for the locally-hosted Gemma 4~31b, a 13.5 percentage-point gap. This gap is not uniform across problem families: on diffusion, transient heat conduction, and solid mechanics, Gemma 4~31b trails GPT~5.2 by only 4--8 points, and on reactor mesh generation it slightly exceeds GPT~5.2 (88\% vs.\ 84\%), indicating that the harness's retrieval and validation loop closes the gap regardless of backend. The largest remaining differences appear in plasticity (60\% vs.\ 84\%), porous flow (68\% vs.\ 92\%), Navier--Stokes (64\% vs.\ 84\%), and phase field (68\% vs.\ 84\%).

Both harness-enabled configurations  outperform their standalone counterparts. Without \gls{moosenger}, the unaided GPT~5.2 API resolves only 10 of 200 prompts (5\%), and Gemma 4~31b standalone fails to resolve any of the 200 prompts (0\%). 

Harness-enabled Gemma 4~31b already reaches 85\% of harness-enabled GPT~5.2's pass rate (76.5\% vs.\ 90\%), whereas removing the harness collapses both models to near-total failure. This shows that the overwhelming majority of end-to-end capability comes from the harness itself, retrieval grounding, validation, and iterative repair, rather than from the choice of backend model.

The residual gap between Gemma 4~31b and GPT~5.2 within the harness, concentrated in plasticity, porous flow, Navier--Stokes, and phase field, suggests that further enhancement of the open-weight configuration, family-specific retrieval augmentation, additional skills and instructions, would likely narrow or close this remaining difference.

\section{Conclusions}
This work extended MOOSEnger, a MOOSE-specialized agent, with a harness that adds grounding in the MOOSE repository, validation and diagnostics, and persistent memory. Across a 200-prompt benchmark, this harness closed most of the performance gap between a locally-hosted Gemma~4 31b model and GPT~5.2. With the harness, Gemma~4 reached 76.5\% success rate and GPT~5.2 reached 90\%, compared to just 5\% and 0\% for their unaided baselines.

In several categories, including diffusion, transient heat conduction, and solid mechanics, the two models performed close to on par. This indicates that the harness, rather than the underlying model, drives most of the capability gain. A larger gap remained in plasticity, porous flow, Navier--Stokes, and phase field, where Gemma~4 trailed GPT~5.2 by 16--24 points, suggesting these categories depended on the base model's reasoning. 

These results suggest that a well-designed harness can make up for much of the gap between open-weight and proprietary models on specialized simulation tasks, a useful finding where export control rule out external API-based models. Closing the remaining gap in the harder categories is a natural next step, through category-specific retrieval, expanded skills, and refined instructions, moving toward fully capable, locally-hosted agents for MOOSE-based multiphysics modeling.

\bibliographystyle{ans}
\bibliography{bibliography}

\end{document}